\documentclass[conference]{IEEEtran}
\usepackage{times}

\usepackage[numbers]{natbib}
\usepackage{multicol}
\usepackage[bookmarks=true]{hyperref}

\usepackage{graphicx}
\usepackage{booktabs}
\usepackage{caption}
\usepackage{subcaption}
\usepackage{float}
\usepackage{xcolor}
\usepackage{amssymb}
\usepackage{amsmath}
\usepackage{algorithm}
\usepackage{algpseudocode}
\usepackage{multirow}

\usepackage{array}

\algnotext{EndIf}% Remove "end if" text
\algnotext{EndFor}% Remove "end for" text
\algnotext{EndWhile}% Remove "end while" text
\algnotext{EndFunction}% Remove "end function" text

\definecolor{light-gray}{gray}{0.95}

\renewcommand\fbox{\fcolorbox{light-gray}{bg-gray}}

\newcommand{\promtblock}[1]{\vspace{1pt}\noindent\fbox{\parbox{0.99\linewidth}{{\fontsize{9}{10} \selectfont  \textcolor{black} {{#1}}}}}\vspace{1pt}}

\newcommand{\llmoutput}[1]{\fcolorbox{highlight}{highlight}{\parbox{0.98\linewidth}{{\fontsize{9}{10} \selectfont  \textcolor{black}{{#1}}}}}}

\newcommand{\blank}[1]{\textcolor{dark-yellow}{#1}}

\definecolor{dark-gray}{HTML}{A9A9A9} % light gray 1
\definecolor{light-gray}{HTML}{b7b7b7} % light gray 1
\definecolor{bg-gray}{HTML}{F8F8F8} % light gray 1
	
\definecolor{green}{HTML}{6aa84f} % green 1
\definecolor{highlight}{HTML}{cfe2f3} % light blue 1
\definecolor{blue}{HTML}{0000ff} % blue 1
\definecolor{brown}{HTML}{9F8C76}
\definecolor{pink}{HTML}{D88782}
\definecolor{light-blue}{HTML}{4a86e8}
\definecolor{dark-yellow}{HTML}{F6AE2D}

\begin{document}

\title{Language-Augmented Symbolic Planner for Open-World Task Planning}

% You will get a Paper-ID when submitting a pdf file to the conference system
% \author{Paper-ID: 424}

% avoiding spaces at the end of the author lines is not a problem with
% conference papers because we don't use \thanks or \IEEEmembership

\author{\authorblockN{Guanqi Chen\authorrefmark{1},
Lei Yang\authorrefmark{1},
Ruixing Jia\authorrefmark{1}, 
Zhe Hu\authorrefmark{2}, 
Yizhou Chen\authorrefmark{3},
Wei Zhang\authorrefmark{4}, 
Wenping Wang\authorrefmark{5},
and Jia Pan\authorrefmark{1}\authorrefmark{3}}
\authorblockA{\authorrefmark{1}The University of Hong Kong}
\authorblockA{\authorrefmark{2}Chongqing Institute of Microelectronics Industry Technology, University of Electronic Science and Technology of China}
\authorblockA{\authorrefmark{3}Centre for Transformative Garment Production (TransGP)}
\authorblockA{\authorrefmark{4}Southern University of Science and Technology}
\authorblockA{\authorrefmark{5}Texas A\&M University}}

\maketitle

\begin{abstract}
Enabling robotic agents to perform complex long-horizon tasks has been a long-standing goal in robotics and artificial intelligence (AI). Despite the potential shown by large language models (LLMs), their planning capabilities remain limited to short-horizon tasks and they are unable to replace the symbolic planning approach. Symbolic planners, on the other hand, may encounter execution errors due to their common assumption of complete domain knowledge which is hard to manually prepare for an open-world setting. In this paper, we introduce a Language-Augmented Symbolic Planner (LASP) that integrates pre-trained LLMs to enable conventional symbolic planners to operate in an open-world environment where only incomplete knowledge of action preconditions, objects, and properties is initially available. In case of execution errors, LASP can utilize the LLM to diagnose the cause of the error based on the observation and interact with the environment to incrementally build up its knowledge base necessary for accomplishing the given tasks. Experiments demonstrate that LASP is proficient in solving planning problems in the open-world setting, performing well even in situations where there are multiple gaps in the knowledge.

\end{abstract}

\IEEEpeerreviewmaketitle

\section{Introduction}

Enabling robotic agents to perform complex long-horizon tasks has been a long-standing goal in robotics and artificial intelligence (AI). To accomplish long-horizon tasks, it is crucial to decompose such a task into a sequence of proper actions with preconditions to ensure that the actions can be correctly executed by the robotic agent, and with effects to ensure that each of the subgoals and eventually the final goal can be reached.

The emergence of Large Language Models (LLMs) in various natural language understanding and instruction following tasks \cite{ouyang2022training,mishra2022cross,sanh2022multitask} motivates the use of LLMs in robotic task planning, aiming to exploit the commonsense reasoning capability of LLMs. Recent studies have demonstrated the capabilities of LLMs in task planning \cite{huang2022language,ahn2022can,huang2022inner,liang2023code,singh2023progprompt}. 
For example, \cite{ahn2022can,huang2022inner} proposed to prompt LLMs to directly synthesize robotic actions based on robot capabilities and adapt to the objects available in the environment. 
On the other hand, some works \cite{liang2023code,singh2023progprompt} provide LLMs with code-style prompts that contain task information, hints, and few-shot examples to synthesize executable codes for robots. 
Despite these endeavors in task planning, growing evidence \cite{valmeekam2023planbench} suggests that achieving long-horizon tasks still poses challenges to LLMs.
A recent work, Demo2Code~\cite{wang2023demo2code}, presents textual demonstrations as supplementary information to generate code for controlling robots in long-horizon tasks. However, the textual demonstrations must be detailed and complete to accomplish the task.

\begin{figure}[t]
\centering
\includegraphics[width=0.48\textwidth]{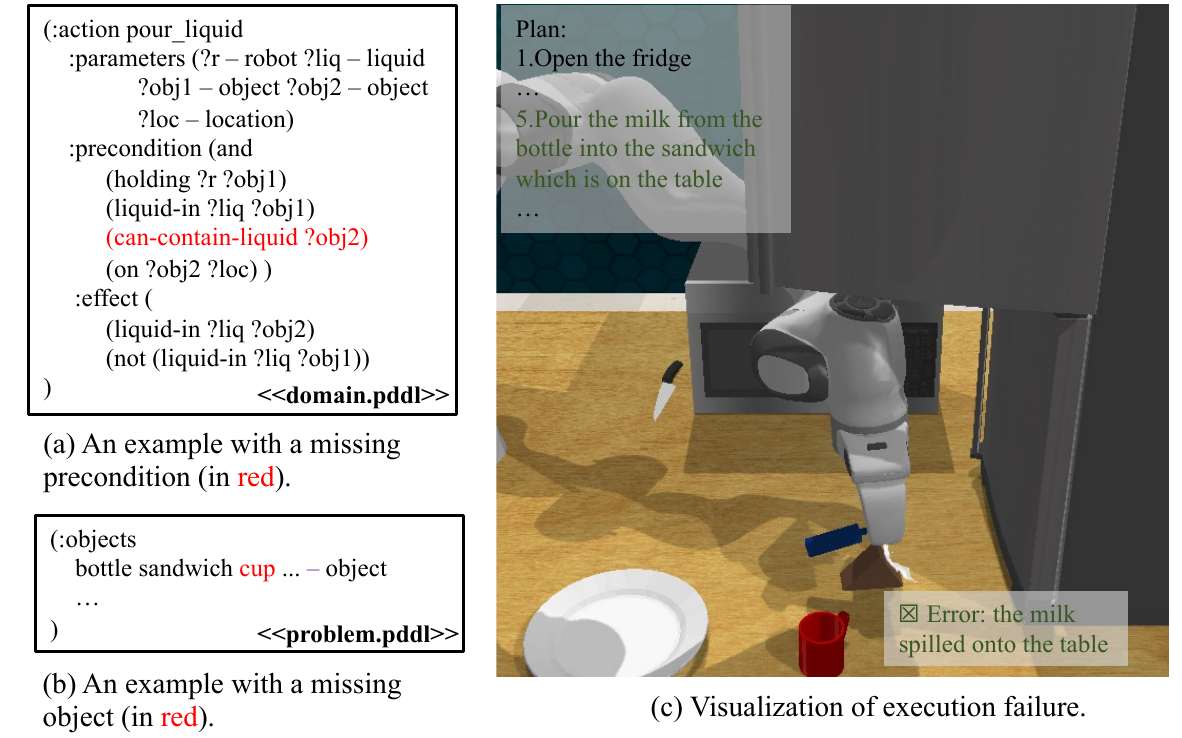}
\caption{An illustrative example of the challenge in open-world task planning is that the knowledge (i.e., domain and problem files shown in (a) and (b)) of the symbolic planner can be incomplete. This causes the execution failure of the planned action as depicted in (c).} \label{fig1}
\end{figure}

Planning with symbolic representations is a systematic approach to synthesizing verifiable plans against a set of specifications of the environment for robotic agents. However, hand-crafted symbolic representations may not be able to capture all constraints of a planning problem, resulting in unexpected failures during execution. 
A few changes in the environment may require a major rewrite of the symbolic plans, making it a tedious process for controlling a robot.
For example, in Fig. \ref{fig1}, consider a scenario where the robot is assigned the task of \textit{heating milk}. The milk is initially in a bottle that is not microwave-safe, therefore it needs to be transferred to another microwave-safe container for heating. But the action \textit{pour\_liquid} may lack a precondition ``\textit{the target object should be able to contain liquid}'', resulting in the robot pouring the milk from the bottle to any object in the scene, or a sandwich in this example as shown in Fig.~\ref{fig1}. This action led to an unexpected effect ``\textit{the milk spilled onto the table}'', which is recognized as an execution failure. This example showcases that the risk of incomplete knowledge about the deployment environment hinders applications of symbolic planning in \textit{open-world} environments where the objects of interest may change and novel preconditions may be necessary to accomplish the same goal.

In this paper, we propose a Language-Augmented Symbolic Planner or LASP. This framework integrates pre-trained large language models to enable conventional symbolic planners to operate in an open-world environment where only incomplete knowledge of action preconditions, objects, and properties is initially available. Like humans, LASP will interact with the environment to collect information and incrementally build up its knowledge base necessary for accomplishing the given tasks. When an execution plan encounters an error alongside the observation of this error, LASP can utilize the LLM to diagnose the cause of the error based on the observation and identify the missing element in the current plan. Recall the example in Fig. \ref{fig1}, LASP can reason the target of the pouring action must be a container based on the observation that the milk spills onto the table. Based on this identified precondition and the goal of the task, LASP can select suitable objects in the environment to expand its knowledge base.

This brings a few clear advantages to LASP. First, LASP keeps the advantages of interpretability, verifiability, and composability that symbolic planning possesses. 
In addition, the observation-based reasoning mechanism enables LASP to correct errors encountered during plan execution. Unlike COWP~\cite{ding2023integrating} that relies on the provided factors that may cause errors, our LASP is designed to automatically identify the potential causes of the errors based on the observations.

In summary, the contributions of this work are as follows:
\begin{enumerate}
    \item We introduce \textbf{LASP}, a novel framework that integrates pre-trained large language models to enable conventional symbolic planners to operate in an open-world environment where only incomplete knowledge of action preconditions, objects, and properties is initially available.
    \item Experiments demonstrate that LASP is proficient in solving planning problems in the open-world setting, performing well even in situations where there are multiple gaps in the knowledge.
\end{enumerate}

\section{Related Work}

\subsection{Task Planning}

\noindent \textbf{Classical Task Planning.}
Task planning aims to determine the sequence of actions that a robot should perform to achieve a specific goal. Conventional methods transfer this planning problem to a search problem via symbolic planners~\cite{helmert2006fast,hoffmann2003metric}. They define the planning problem in a declarative language, such as planning domain definition language (PDDL)~\cite{aeronautiques1998pddl,fox2003pddl2}, specifying the initial state, goal, and a set of actions that the robot can take to transition between states. The algorithms then search through the space of possible action sequences to find a plan that starts from the initial state and reaches the given goal. However, the symbolic planner relies on complete domain knowledge which encompasses all the constraints of the environment to search for a valid solution. This is hard and tedious for a non-expert to tailor. 
To address this challenge, Lin et al. \cite{lin2021change, lin2023towards} proposed an approach to repair flawed action preconditions and effects in the planning models to make the plans viable, focusing on scenarios where the planning models contradict the provided valid plans.
Gragera et al. \cite{gragera2023planning} proposed an automated planning approach to repair planning models with incomplete action effects, allowing symbolic planners to generate solutions for tasks that were originally unsolvable. Besides, Sreedharan et al. \cite{sreedharan2019can} introduced a method using hierarchical abstractions to generate explanations for unsolvable planning problems. These explanations can be further utilized for the planning model repair to make the planner find a solution. In our work, we focus on enabling symbolic planners to operate in the open-world environment where only incomplete knowledge of action preconditions, objects, and properties is initially available. When planning with such incomplete knowledge, the planner may generate an invalid plan that leads to execution errors. Our proposed LASP can leverage LLMs to repair the planning model based on the textual observation of the execution error, enabling the planner to find a valid solution. Compared to the works by Liu et al. \cite{lin2021change, lin2023towards}, which repair the planning model to make the provided valid plan viable, our proposed LASP starts with an incorrect plan, refining the planning model to find a valid plan. In contrast to the work by Gragera et al. \cite{gragera2023planning} which can repair missing action effects, LASP is able to repair missing action preconditions, properties, and objects. Unlike the work by Sreedharan et al. \cite{sreedharan2019can}, our work mainly focuses on repairing the incomplete planning model and enabling the symbolic planner to find solutions rather than explanations. Besides, our method does not conflict with these works and can be used in conjunction with them.

\noindent \textbf{Planning with LLMs.}
Recently, after training on massive data, LLMs have acquired rich commonsense knowledge and demonstrated powerful in-context learning ability. Consequently, numerous researchers are exploring applications of these models in task planning. Huang et al.~\cite{huang2022language} attempted to query LLMs to predict actions for task completion with several demonstrations and the task description in natural language. Ahn et al.~\cite{ahn2022can} proposed SayCan to ground the free-form output of LLMs in robotic affordances. Some works \cite{liang2023code,singh2023progprompt} provided code-style prompts to LLMs and required LLMs to generate codes that the robot can execute directly. However, a significant drawback of existing LLMs is that they exhibit limited proficiency in undertaking long-horizon planning endeavors for complex tasks~\cite{valmeekam2023planbench}. 
Wang et al.~\cite{wang2023demo2code} introduced Demo2Code to leverage textual demonstrations as supplementary information to generate code for controlling robots in long-horizon tasks. However, Demo2Code relies on detailed and complete textual descriptions of completing the task. Therefore, some researchers \cite{liu2023llm+,xie2023translating} explored using LLMs to translate the task description from natural language to PDDL and leveraged the symbolic planner to solve the planning problem.
In this work, we leverage LLMs to enable symbolic planners to operate in an open-world environment where only incomplete knowledge of action preconditions, objects, and properties is initially available. When the robot encounters an error during executing the plan, we utilize the commonsense reasoning capability of LLMs to identify the cause of the error, and further refine the planning model. Additionally, in instances where the planner cannot find a solution due to the incomplete object set, we harness the commonsense reasoning ability of LLMs to supplement missing objects, thereby facilitating the planner in reaching a solution.

\subsection{Robot Error Correction}

Recovering from failure or correcting an error is a significant capability for robots, especially in scenarios where interactions occur with non-expert users or in the absence of human assistance. Liu et al.~\cite{liu2023reflect} proposed REFLECT, a framework that utilizes LLMs to summarize robot experiences, explain the error, and predict actions to correct the error directly. Huang et al.~\cite{huang2022inner} introduced Inner Monologue that can leverage a variety of sources of feedback from the environment and humans for replanning to predict new actions to continue completing tasks. Compared to the works by Liu et al.~\cite{liu2023reflect} and Huang et al.~\cite{huang2022inner}, our proposed LASP can avoid repeating the same errors after correction. Besides, Ding et al.~\cite{ding2023integrating} designed COWP to identify possible errors in a plan based on the provided situation and utilize LLMs to augment the robot’s action knowledge to find a new plan to correct the possible erroneous actions. In contrast to COWP~\cite{ding2023integrating}, our proposed approach can eliminate the dependence on provided error factors and identify potential sources of errors.

\section{Background and problem statement}
In this work, we employ a PDDL-based symbolic planner to perform task planning. 

\paragraph{Planning Task}
A planning task is defined as a tuple $P = (O, R, s_0, g, A)$, 
consisting of the following components:

\begin{itemize}
\item $O$ is the set of objects in the planner's internal world;
\item $R$ is the set of properties of each object in $O$. %of interest to us regarding the object. 
The property can be a binary-valued predicate or a numeric-valued function. We define $F$ as the set of propositions instantiated from $R$ with respect to $O$, and $2^F$ refers to the state space;
\item $s_0 \in 2^F$ is the initial state. For each time step $t$, we denote $s_t \in 2^F$ as the state of this step, i.e., the set of its true propositions;
\item $g \subseteq F$ is the goal description  composed of a set of propositions;
\item $A$ is the set of actions that can change the current state by adding or deleting some propositions. Each action can be specified by a triplet $a = ( \textrm{Pre}(a), \textrm{Add}(a), \textrm{Del}(a))$, 
where $\textrm{Pre}(a), \textrm{Add}(a), \textrm{Del}(a) \subseteq F$ represent the set of preconditions, the set of added propositions, and the set of removed propositions that modify the current state. If an action $a_t$ is applicable in a state $s_{t-1}$, it implies $\textrm{Pre}(a) \subseteq s_{t-1}$. And the resulting state $s_t$ should become $(s_{t-1} \setminus \textrm{Del}(a)) \cup \textrm{Add}(a)$.
\end{itemize}

The solution to $P$ is a plan consisting of a sequence of actions $\pi = (a_1, a_2, ..., a_n)$ that can achieve the goal $g$ starting from $s_0$.

\paragraph{Types of Incomplete Knowledge}
In our open-world setting, we assume that $O$, $R$, and $\textrm{Pre}(a)$ may be incomplete. Here, we denote $\bar{O}$, $\bar{R}$, $\bar{A}$ as complete object, property, and action sets.
\begin{itemize}
\item \textbf{Incomplete} $\textrm{Pre}(a)$. In this case, the state $s_{t-1}$ before executing action $a_t$ may not satisfy the actual preconditions $\textrm{Pre}(\overline{a_t})$, where $\bar{a_t} \in \bar{A}$. This can lead to the agent's failure to execute the action $a_t$ to obtain the desired result. For example, when the precondition ``\textit{the target object should be able to contain liquid}" is missing in the precondition set of action \textit{pour\_liquid}, the planner may instantiate an action of pouring milk from the bottle to the sandwich, resulting in an execution error. 

\item \textbf{Incomplete} $R$.
Without a complete set of properties, the planning task may lack some propositions, for example, ``\textit{the target object is a liquid container}''. This may lead to incomplete $\textrm{Pre}(a)$, eventually leading to the previous error type. 

\item \textbf{Incomplete} $O$. Lastly, the lack of a proper object prevents finding a solution to $P$. For example, when no object in $O$ is simultaneously \textit{microwave-safe} and \textit{able to contain liquid}, the planner will not find a solution to the planning task shown in the \textit{heating milk} example.  
\end{itemize}

In this work, we focus on refining the aforementioned incomplete knowledge within the planning task $P$ to find a plan that can achieve the given goal $g$.

We assume that there exists a valid solution for the planning problem with complete domain knowledge. To verify whether an action can be successfully executed in the environment, we introduce an oracle that contains the complete domain knowledge for this purpose. When the state of the environment aligns with the preconditions specified in the oracle PDDL model for a particular action, it can be successfully executed. Conversely, if the preconditions are not met, the agent will be given a few sentences describing the observation $d(e)$ of the error $e$ during execution failure. In the practical applications, vision-and-language models\cite{radford2021learning, li2022blip, gpt4v} can be used to generate a natural language paragraph describing images that capture the scene during execution failure, serving as $d(e)$.Given $d(e)$, the agent needs to infer the cause of the error through interactions with the environment to refine the planning task, updating the precondition set for the actions it takes, expanding its object set $O$ with elements in $\overline{O}$, and acquiring novel properties of an object to enlarge its property set $R$ from $\overline{R}$.

\section{Methodology}

\begin{algorithm}
\caption{LASP: Language Augmented Symbolic Planning}
\label{alg-lasp}
\begin{algorithmic}[1]
\Require
Planning Problem $P = (O, R, s_0, g, A)$;
Symbolic Planner SP;
Large Language Model LLM.
\Ensure Plan $\boldsymbol{\pi}$. 

\Statex \textcolor{green}{// Refine object set $O$}
\Function{refine\_object\_set}{$P'$, $d(e)$}
    \State $\text{objects} = \textsc{get\_candidate\_objects\_from\_env}()$
    \State $\text{new\_object} = \textsc{LLM}(P', \text{objects},d(e))$
    \State $P' = \textsc{add\_knowledge}(P', \text{new\_object})$
    \State \Return $P'$
\EndFunction

\Statex \textcolor{green}{// Refine precondition set $Pre(a)$ and property set $R$}
\Function{refine\_prec\_and\_prop\_set}{$P$, reason, d(e)}
    \State $\text{new\_prec}, \text{new\_prop} = \textsc{LLM}(P, \text{reason}, d(e))$
    \State $P' = \textsc{add\_knowledge}(P, \text{new\_prec}, \text{new\_prop})$
    \State \Return $P'$
\EndFunction

\Statex \textcolor{green}{// The refining step}
\Function{refining\_step}{$P$, $\text{reason}$, $d(e)$}
    \State $P' = \textsc{refine\_prec\_and\_prop\_set}(P, \text{reason}, d(e))$
    \State $\pi' = \textsc{SP}(P')$
    \If {$\pi'$ is empty}
        \State $P' = \textsc{refine\_object\_set}( P', d(e))$
    \EndIf
    \State \Return $P'$
\EndFunction

\Statex \textcolor{green}{// Refine and replan recursively}
\Function{refine\_and\_replan}{$P$, $\pi$, $\text{done}$, $d(e)$}
    \If{$\text{done}$}
        \State \Return $P$, $\pi$, $\text{done}$, $d(e)$ \textcolor{green}{// Stop the recursion}
    \EndIf

    \State $\text{reasons} = \textsc{LLM}(d(e), \pi)$
    \For{$\text{reason}$ in $\text{reasons}$}
        \State $ P' = \textsc{refining\_step}(P, \text{reason}, d(e))$ 
        \State $\pi' = \textsc{SP}(P')$
        \If{ $\pi'$ is empty}
            \State \textbf{continue}
        \EndIf
        \State $\text{done}', d(e') = \textsc{execute}(\pi')$
        \If{$\text{done}'$}
            \Statex \textcolor{green}{$\qquad\quad$// 1) Robot accomplishes the task}
            \State \Return $P'$, $\pi'$, $\text{done}'$, $d(e')$
        \Else
            \If{$d(e') \neq d(e)$}
                \Statex \textcolor{green}{$\qquad\qquad\quad$// 2) New error encountered}
                \State $ P'', \pi'', \text{done}'', d(e'')$ \Statex $\qquad\qquad\qquad = \textsc{refine\_and\_replan}(P', \pi', \text{done}', d(e'))$ 
                \If{$\text{done}''$}
                    \State \Return $P''$, $\pi''$, $\text{done}''$, $d(e'')$
                \EndIf
            \Else
                \Statex \textcolor{green}{$\qquad\qquad\quad$// 3) Error not corrected}
                \State \textbf{continue} 
            \EndIf
        \EndIf
    \EndFor
    \Statex \textcolor{green}{$\qquad$// Error not corrected after all the attempts}
    \State \Return $P$, $\pi$, $\text{done}$, $d(e)$   

\EndFunction

\Function{main}{ }
    \State $\pi = \textsc{SP}(P)$
    \State $\text{done}, d(e) = \textsc{execute}(\pi)$
    
    \If{not $\text{done}$}
        \State $ *, \pi, *, * = \textsc{refine\_and\_replan}(P, \pi, \text{done}, d(e))$
    \EndIf
    \State \Return $\pi$
    
\EndFunction
\end{algorithmic}
\end{algorithm}

\begin{figure*}[t]
\centering
\includegraphics[width=0.98\textwidth]{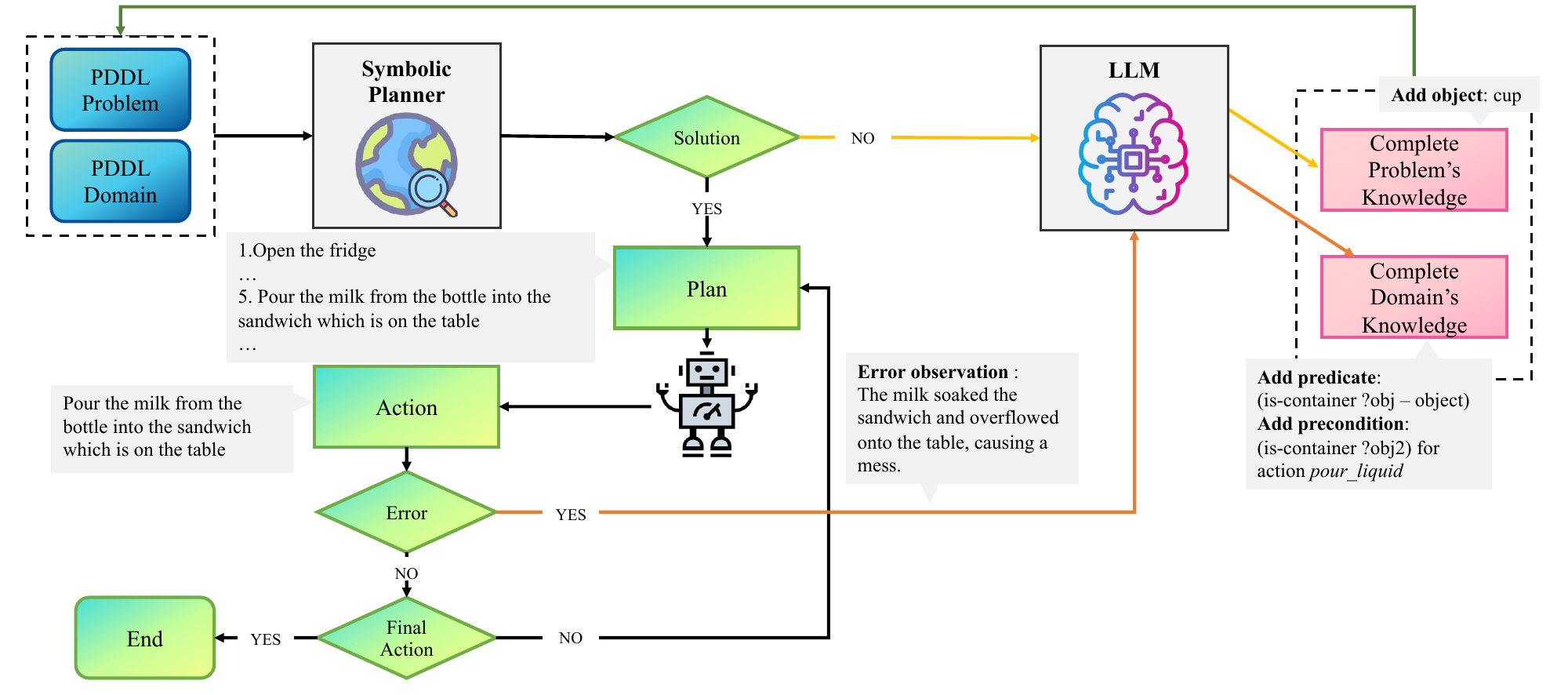}
\caption{An overview of our proposed LASP, a task planning framework in the open world. The symbolic planner is responsible for finding a plan to accomplish the given task, and subsequently, the robot executes this plan. 
Given an initial, incorrect plan, the robot will encounter some errors during action execution. Our proposed method LASP will invoke the LLM to recursively refine the task planning problem by supplementing action knowledge to the planner. Moreover, if the planner is unable to find a plan afterward, the LLM can augment the planner's task-specific knowledge with necessary objects to assist the planner in finding an error-free solution.} \label{fig_pipeline}
\end{figure*}

LASP enables symbolic planners to solve planning problems in an open-world setting with the support of LLMs. The overview of LASP is shown in Fig. \ref{fig_pipeline}. 
While the symbolic planner can find a solution to a given task if the problem $P$ is complete, in the case of an incomplete problem, our LLM is tasked to refine and complete the problem based on the error observations through interactions with the environment.
In the following section, we first introduce the proposed algorithm and then elaborate on the functionalities of our LLM.

\subsection{LASP Algorithm}
As shown in Algorithm \ref{alg-lasp}, the symbolic planner initially generates a plan $\pi$ for the given planning problem $P$ (Line 37). The robot then interacts with the environment by sequentially executing the actions in the plan.
In cases of an execution error, LASP is provided with a description of the observation regarding the error ($d(e)$) and then it invokes a recursive function \textsc{refine\_and\_replan} to refine the planning problem according to the error observation $d(e)$. 

In the step of refining the planning problem, LASP first utilizes the LLM to infer potential reasons causing the error (Line 19). 
Based on the list of potential reasons, LASP employs the LLM to progressively generate a precondition regarding the action, i.e., \textsc{refine\_prec\_and\_prop\_set}, and add it to the planner. If the existing properties are insufficient to express the precondition, LASP also will add corresponding new properties to the planner. 
However, adding a precondition does not guarantee a solution due to the lack of a suitable object. Hence, LASP then tries to search for a suitable object from $\overline{O}$ using the LLM to augment the planner's knowledge by using the function \textsc{refine\_object\_set}. 

After each refining step, the symbolic planner undergoes a replanning step. If a new plan $\pi'$ is found, it will be executed to check whether the goal is reached. There are three possible outcomes: (1) the robot accomplishes the task; (2) the robot encounters a new error observation (i.e., $d(e') \neq d(e)$, where $e'$ and $e$ refer to the new error and the previous one, respectively, that are latent and not observable); (3) the error is not corrected. 
For the second situation, LASP will recursively continue the refining and replanning step. For the last, LASP will iterate over all potential reasons to refine the problem.

\subsection{Functionalities of LLM}
In our proposed algorithm, the LLM plays five roles: (1) cause analyzer, (2) action precondition generator, (3) property completeness evaluator, (4) NL-to-PDDL translator, and (5) object expander. The first four modules (or functions) are designed to correct errors during action execution, while the last function is intended for scenarios where the planner cannot find a solution due to the lack of objects.

In the following, we give an example to walk the reader through these LLM-empowered functions. We denote the variables in a template prompt as \textsc{[\$VARIABLES\$]} and we provide examples of actual textual descriptions for some variables. %due to limited space.

\textbf{Cause Analyzer. }
The cause analyzer is designed to generate potential causes of the execution error. We provide a rich context to the LLM, allowing it to have a clear understanding of the robot's objective, what it has done, the incorrect action it has taken, and the corresponding observation. This enables the LLM to make reasonable speculations about the causes of errors.

\promtblock{\underline{\textbf{Prompt:}} You are expected to provide possible reasons for errors in robot actions, where errors are essentially unexpected outcomes. You are given the robot's task, the historical actions, the action that caused the error, and the observation of the error. 
\textcolor{brown}{[\$OUTPUT REQUIREMENTS\$]}
\newline
Given \blank{[\$ROBOT TASK\$]}: The robot's task is to heat the milk and place it on the table.}
\promtblock{\blank{[\$HISTORICAL ACTIONS\$]}: The historical actions are:\\
1. Open the fridge.\\
2. Pick up the sandwich from the fridge.\\
3. Put the sandwich on the table.\\
4. Pick up the bottle from the fridge.\\
\blank{[\$ERRONEOUS ACTION\$]}: The action caused the error is: Pour the milk from the bottle into the sandwich which is on the table.\\
\blank{[\$ERROR OBSERVATION\$]}: The observation of the error is: The milk soaked the sandwich and overflowed onto the table, causing a mess.
\newline
What are possible reasons that could lead to this error?\\ \\
\llmoutput{\underline{\textbf{LLM output:}} 
Possible reasons:\\
\blank{[\$REASON\$]} 1: Misinterpretation of the task instructions: The robot was supposed to pour the milk into a container for heating, not onto the sandwich.\\
\blank{[\$REASON\$]} 2: Lack of spatial awareness: The robot may not have recognized the presence of the sandwich on the table.\\
...}}

\textbf{Action Precondition Generator. }
With a potential cause, the action precondition generator is tailored to produce a few sentences to describe the precondition of the action to prevent the error. 

\promtblock{\underline{\textbf{Prompt:}}
You are given a robotic action that caused an error and the reason why this error occurred. You are expected to provide a precondition for a robotic action to prevent a recurrence of the same error. \textcolor{brown}{[\$OUTPUT REQUIREMENTS\$]} \newline
Given \blank{[\$ERRONEOUS ACTION\$]} and \blank{[\$REASON\$]}. 
\newline
What precondition should be satisfied to prevent a recurrence of the same error?
\newline
\newline
\llmoutput{\underline{\textbf{LLM output:}} \blank{[\$NL-PRECONDITION\$]} Precondition: The target of the pouring action must be a container.}}

\textbf{Property Completeness Evaluator. }
Due to the possibility of incomplete properties in the planner, the property completeness evaluator is developed to determine whether properties are missing. Specifically, the LLM will assess whether the existing properties adequately represent the precondition obtained in the action precondition generator.

\promtblock{\underline{\textbf{Prompt:}}
You are expected to judge whether the given natural language described precondition can be expressed by the candidate predicates and functions, or their respective antonyms. You are given the natural language described precondition, candidate predicates, and functions. \textcolor{brown}{[\$OUTPUT REQUIREMENTS\$]} \newline
Given \blank{[\$NL-PRECONDITION\$]}\\
\blank{[\$CANDIDATE PREDICATES\$]}: (holding ?r - robot ?obj - object), (is-empty-handed ?r - robot), (is-open ?rec - receptacle), (in ?obj - object ?rec - receptacle), (on ?obj - object ?loc - location), (is-microwave ?rec - receptacle), (is-heat-insulation ?obj - object), (liquid-in ?liq - liquid ?obj - object), (is-microwave-safe ?obj - object), (can-support ?loc - location).\\
\blank{[\$CANDIDATE FUNCTIONS\$]}: (temperature ?obj - object), (tolerance-temperature ?r - robot), (liquid-temperature ?liq - liquid). \\
Can the precondition be expressed by the candidate predicates and functions, or their respective antonyms?
\newline
\newline
\llmoutput{\underline{\textbf{LLM output:}} No}}

\textbf{NL-to-PDDL Translator. } 
Depending on the output of the property completeness evaluator, the NL-to-PDDL translator has two ways to translate the natural language described precondition into PDDL expression. If the property completeness evaluator determines that the existing properties cannot express the precondition, the NL-to-PDDL translator can convert the precondition without requiring the translated PDDL precondition related to any existing properties. Otherwise, a constraint is imposed to translate the precondition based on the existing properties (included in the output of the property completeness evaluator).

\promtblock{\underline{\textbf{Prompt:}}
Given the action description, the candidate parameters of the action, the corresponding between instances and parameters, and a natural utterance, you are expected to translate the natural utterance into a PDDL precondition expression for the given action. \textcolor{blue}{[\$PDDL GRAMMARS\$]} \textcolor{brown}{[\$OUTPUT REQUIREMENTS\$]} and \textcolor{light-blue}{[\$TWO EXAMPLES\$]} \newline
Given \blank{[\$ERRONEOUS ACTION\$]} \\
\blank{[\$ACTION PARAMETERS\$]}: Candidate parameters: the robot, denoted as ?r; the liquid, denoted as ?liq; the object, denoted as ?obj1; the object, denoted as ?obj2; the location, denoted as ?loc.\\
\blank{[\$CORRESPONDENCE\$]}: Corresponding between instances and parameters: the agent is ?r; the milk is ?liq; the bottle is ?obj1; the sandwich is ?obj2; the table is ?loc.\\
\blank{[\$NL PRECONDITION\$]}.
\newline
\newline
\llmoutput{\underline{\textbf{LLM output:}} PDDL expression: (is-container ?obj2)}}

When a new property is added to the planner, the planner needs to acquire the values of every existing object regarding this newly added property for planning. To accomplish this, we assess the similarity between the newly added property and properties in the environment, except those already present in the planning problem. Subsequently, we assign the value of the most semantically similar property to this new property. This enables the planner to incorporate values for every existing object concerning this newly added property into the initial state.

\textbf{Object Expander.}
When a new precondition is introduced into the symbolic planner, the existing objects in the symbolic planner may no longer be sufficient to meet the new constraints. Hence, the planner may fail to find a solution. To address this problem, the object expander is introduced to add one or more new objects to the planning problem, thereby allowing the planner to find a solution. Here, we query the object expander to select a new object to assist the robot complete the task.

\promtblock{\underline{\textbf{Prompt:}}
You are expected to select the most suitable object from the candidate objects to prevent a recurrence of the error and assist the robot complete the task. You are given the robot task, the action that causes the error, the precondition that the action does not satisfy, and the candidate objects. \textcolor{brown}{[\$OUTPUT REQUIREMENTS\$]}\newline
Given \blank{[\$ROBOT TASK\$]} \blank{[\$ERRONEOUS ACTION\$]} \blank{[\$NL PRECONDITION\$]}\\
\blank{[\$CANDIDATE OBJECTS FROM ENV\$]}: cup, plate, knife.}
\promtblock{Which object is most suitable to prevent a recurrence of the error and assist the robot complete the task?
\newline
\newline
\llmoutput{\underline{\textbf{LLM output:}} Suitable object: cup.}}

\section{Experiments}

\subsection{Experimental Setup}

\paragraph{Evaluation tasks} In our experiments, we designed 4 tasks in the kitchen domain to evaluate the proposed method LASP: \textit{serving fruit}, \textit{storing fruit}, \textit{heating sandwich}, and \textit{heating milk}. For each task, we constructed at least a pair of incomplete PDDL domain and problem files to test LASP. We constructed a total of 9 pairs of incomplete PDDL domain and problem files. Among the 9 test cases, a crucial precondition is missing in seven domain files and two crucial preconditions are missing in each of the other two domain files. More information about the missing knowledge of test cases can be found in Tab.~\ref{lasp_test_cases}.

\begin{table*}[t]
    \centering
    \begin{tabular}{m{1.4cm}| m{1.8cm} m{6cm} m{3cm} m{2cm}}
    \toprule
    \textbf{Test case} & \textbf{Task}  &  \textbf{Missing preconditions} & \textbf{Missing properties} & \textbf{Missing objects} \\
    \hline
    1 & serve fruit & Picking up an object requires its weight to be below the robot's lifting capacity. & weight and lift-capacity & apple \\
    2 & store fruit & Placing an object in the fridge is conditioned upon the door being open. & -  & - \\
    3 & heat sandwich & Picking up an object requires its temperature to be below the robot's tolerance temperature. & temperature and tolerance-temperature & glove \\
    4 & heat sandwich & Using a microwave to heat objects is conditioned upon the door being closed. & - & - \\
    5 & heat milk & Pouring liquid from object1 to object2 requires object2 can contain liquid.  & able to contain liquid & cup \\
    6 & heat milk & Wearing the heat-insulator requires the robot not holding anything. & - & - \\
    7 & heat milk & Picking up an object requires its temperature to be below the robot's tolerance temperature. & temperature and tolerance-temperature & glove \\
    \hline
    \multirow{2}{*}{8} & \multirow{2}{*}{heat sandwich} & \multicolumn{3}{c}{\multirow{2}{*}{Combination of Case 3 and 4}} \\
    ~ & ~ & \multicolumn{3}{c}{~} \\
    \multirow{2}{*}{9} & \multirow{2}{*}{heat milk} & \multicolumn{3}{c}{\multirow{2}{*}{Combination of Case 5 and 7}} \\
    ~ & ~ & \multicolumn{3}{c}{~} \\
    \bottomrule
    \end{tabular}
    \caption{Test cases and their missing knowledge for evaluating LASP.}
    \label{lasp_test_cases}
\end{table*}

\begin{table*}[h]
    \centering
    \begin{tabular}{c|c|c|c|c|c}
    \toprule
    % \hline
        Task & Optimal Plan Steps & LMZSP~\cite{huang2022language} & Inner Monologue~\cite{huang2022inner} & Progprompt~\cite{singh2023progprompt} & LASP (ours)  \\
        \hline
        serve fruit & 2  & 20\% & 100\% & 80\% & 100\% \\
        store fruit & 3  & 20\% & 100\% & 100\% & 100\% \\
        \hline
        heat sandwich & 10 & 0\% & 20\% & 0\% & 100\% \\
        heat milk & 13 & 0\% & 20\% & 0\% & 85\% \\
    \bottomrule
    % \hline
    \end{tabular}
    \caption{Quantitative results of LASP and the language-driven methods in open-world task planning. Our method can achieve superior results as compared to the language-driven planning methods}
    \label{tab_results}
\end{table*}

\begin{table}[h]
    \centering
    \begin{tabular}{c|c|c}
    \toprule
        Test case & Task & LASP (ours) \\
        \hline
        1 & serve fruit& 100\% \\
        2 & store fruit& 100\% \\
        \hline
        3 & heat sandwich& 100\% \\
        4 & heat sandwich& 100\% \\
        5 & heat milk& 80\% \\
        6 & heat milk& 100\% \\
        7 & heat milk& 100\% \\
        \hline
        8 & heat sandwich& 100\% \\
        9 & heat milk& 60\% \\
    \bottomrule
    \end{tabular}
    \caption{Success rate of our approach on test cases.}
    \label{tab:lasp_result}
\end{table}

\paragraph{Baselines}
To demonstrate the advantages of our method, we compare with the following baselines:
\begin{itemize}
    \item Language Models as Zero-Shot Planners (LMZSP)~\cite{huang2022language} is a baseline that queries an LLM to generate a plan for a given task with few-shot examples. In this way, the output of the LLM is not grounded, so it utilizes a pre-trained language model to translate each output action into an available action.
    \item Inner Monologue \cite{huang2022inner} predicts an initial plan based on the prompt that contains the task description, the primitive actions, available objects in the environment, and few-shot examples. Subsequently, it replans in response to feedback from the environment whenever an error is encountered during plan execution. Here, we provide information on the action that encounters an error and the observation of the error to the LLM for its replanning. Two rounds of feedback are allowed.
    \item ProgPrompt \cite{singh2023progprompt} leverages a code-style prompt to query an LLM to generate a code to accomplish the given task. Its prompt includes available action functions, available parameters to call the functions, and few-shot examples.
\end{itemize}

\paragraph{Evaluation metric}
We evaluate the effectiveness of different approaches regarding the success rate. For each task, we execute each baseline method five times and report the average success rate. As for our method, we perform five runs for each case within a task and report the average success rate for that specific task across all cases within the same task.

\paragraph{Implementation details}
% implementation of  generating error observation 
We utilize a numeric fast downward planner\footnote{https://github.com/ipc2023-numeric/team-1} to search plans for given planning tasks. Besides, both the baselines and our approach leverage GPT-4 in the implementation. Additionally, to facilitate providing diverse descriptions of error-related observations, we leverage GPT-4 to generate these descriptions based on erroneous actions, unmet conditions, and reference examples.

\begin{figure*}[t]
\centering
\includegraphics[width=0.98\textwidth]{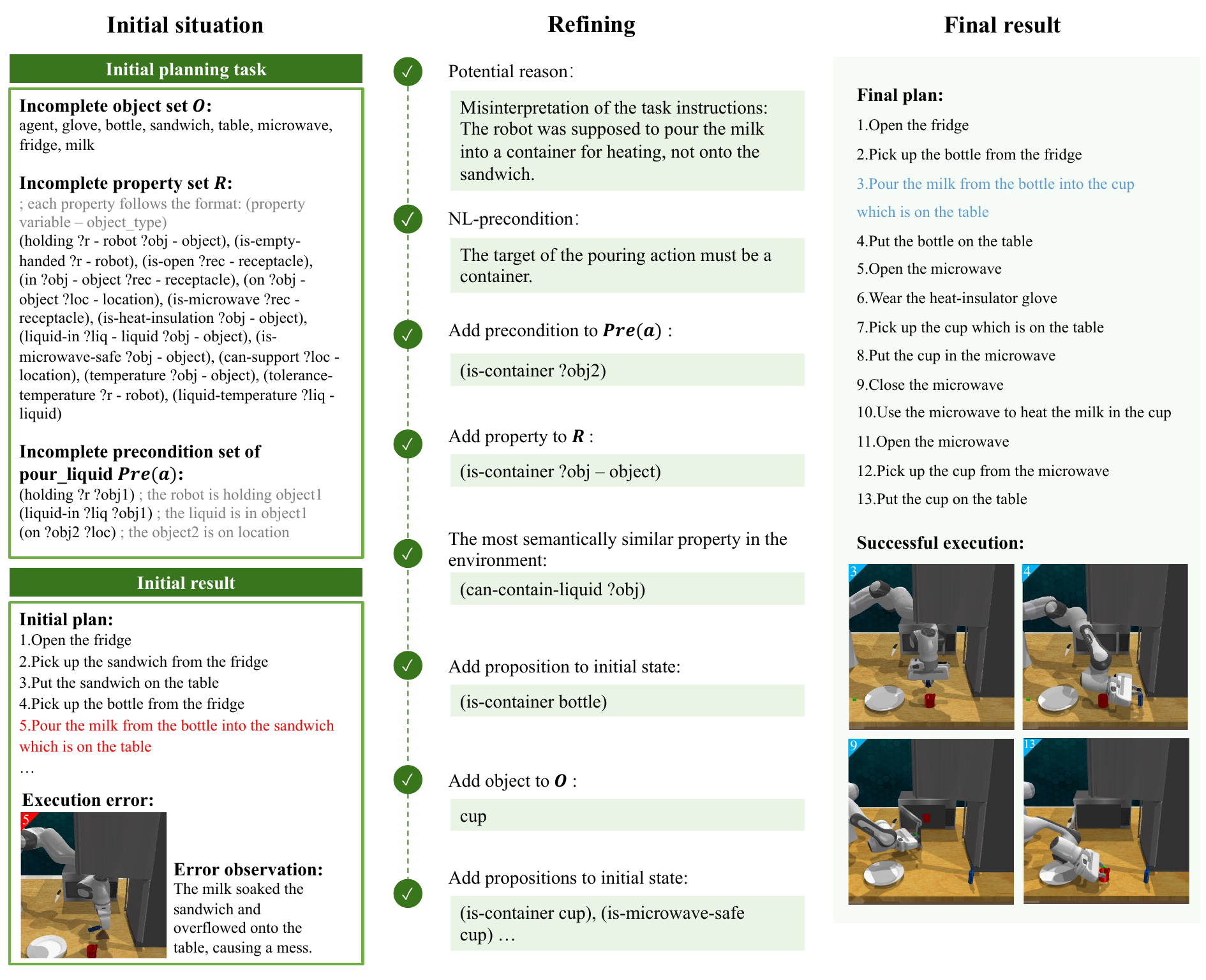}
\caption{Qualitative result of LASP for the task of \textit{heating milk}. LASP is able to refine incomplete knowledge within the planning task and find a plan to accomplish the task. The snapshots of the execution actions are taken from our simulation platform built on PyRep~\cite{james2019pyrep}. Comments are indicated in gray text and start with a semicolon.} \label{fig_visualization}
\end{figure*}

\subsection{Results}

The experimental results are shown in Table~\ref{tab_results}. Among them, our proposed LASP achieves the highest success rate for all the tasks. We could notice that LASP performs well in handling both short-horizon tasks (i.e., \textit{serving fruit} and \textit{storing fruit}) and long-horizon tasks (i.e., \textit{heating sandwich} and \textit{heating milk}). Tab.~\ref{tab:lasp_result} presents the detailed performance of LASP on test cases. LASP achieves 100\% success rates in 7 cases with multiple tests. Even in instances where there are two missing preconditions, as constructed in cases 8 and 9, LASP demonstrates superior performance, achieving an average success rate of $80\%$. For long-horizon tasks such as \textit{heating sandwich} and \textit{heating milk}, LASP attains an average success rate of $91.4\%$ across all test cases related to these long-horizon tasks. In comparison, other purely LLM-driven methods exhibit poor performance in handling long-horizon planning tasks, highlighting the advantage of symbolic planning, which incorporates logical reasoning mechanisms to explicitly reason about the constraints and conditions. 

For \textit{LMZSP}, it is difficult to ground the commonsense knowledge to a specific domain via the few-shot examples, which leads to its poor performance on all the tasks. For example, when the robot was asked to finish the task of storing an apple in a closed fridge, \textit{LMZSP} instructed the robot to first grasp an apple and then attempt to open the fridge. However, it failed to open the fridge because the robot's hand was occupied with holding the apple, preventing it from manipulating the fridge door handle. 
\textit{Inner Monologu}e lacks a clear understanding of the constraints on actions and its own state after each action, leading to frequent omissions of crucial actions or disruptions in the correct sequencing of actions. Therefore, it is difficult to generate a long plan for complex tasks. For example, it instructed the robot to place the cup on the table when the robot did not hold the cup. 
For \textit{Progprompt}, there are some noise and unforeseen conditions in its generated codes, which results in its poor performance for long-horizon planning tasks. For example, it considers the temperature of the sandwich being higher than the tolerance temperature of the robot as a precondition for picking up the sandwich; however, this precondition is incorrect.

The qualitative result is shown in Fig. \ref{fig_visualization}. Starting with the initial planning task which involves incomplete knowledge of the action precondition set, the object set, and the property set, LASP is able to refine the planning task based on the error observation through interactions with the environment, which enables the planner to find a plan to accomplish the task.

\subsection{Discussions and Limitations}

Firstly, the performance of our approach largely depends on the capability of the LLM to read PDDL expressions and translate natural language descriptions into PDDL expressions. If the initial PDDL expressions are poorly readable by the LLM, our approach may suffer a performance drop. The failure cases of \textit{heating milk} are due to this reason. In these failure examples, the action precondition generator has synthesized the natural language described precondition, e.g., ``the target for the milk pouring action is a container" or ``the target vessel must be suitable for containing liquids", indicating \textit{the target of the pouring action should be a container}.
However, the property completeness evaluator falsely inferred these preconditions described in natural language can be expressed using existing properties, leading to the failures. This problem may be solved by translating the PDDL expressions in the prompts into natural language.

Currently, our approach assumes that an environment can provide the robotic agent with the true property values at request. 
In real-world applications, vision-language models (or even humans) may wrongly estimate the property value of an object in the environment.
In the future, we intend to estimate the property values from noisy data using vision-language models and leverage relationships between different properties as well as robotic interactions with the environment to gradually reduce the uncertainty of the estimated values. This would make LASP applicable to real-world scenarios.

Besides, our approach can update the planner's knowledge base only after an action error occurs. In contrast, humans can update their knowledge by reading books (e.g., a manual) or watching demonstrations without explicitly triggering an execution error. Humans can also predict the future states ahead to verify their actions. In the future, we aim to incorporate these capabilities into our method.

Additionally, the assumption that the action effects are complete in our approach may not hold in practical applications. In the future, we are going to refine the incomplete action effects with the help of additional information like demonstrations.

\section{Conclusion}
In this paper, we introduce a novel framework that integrates pre-trained large language models to enable conventional symbolic planners to operate in an open-world environment where only incomplete knowledge of action preconditions, objects, and properties is initially available.
LASP keeps the advantages of interpretability, verifiability, and composability that symbolic planning possesses.  
In addition, the observation-based reasoning mechanism enables LASP to correct errors encountered during plan execution. 
Experiments demonstrate that LASP is proficient in solving planning problems in the open-world setting, performing well even in situations where there are multiple gaps in the knowledge.

\section*{Acknowledgments}
This research project is partially supported by the Innovation and Technology Commission of the HKSAR Government under the InnoHK initiative, Hong Kong General Research Fund (11202119 and 11208718), Innovation and Technology Commission (GHP/126/21GD), and Guangdong, Hong Kong and Macao Joint Innovation Project (2023A0505010016). Jia Pan is the corresponding author. The authors would like to thank Jingqian Liu, Xuebo Ji, and Wentao Chen for their help in drawing and discussions. Also, we would like to thank blind reviewers for their constructive suggestions.

%% Use plainnat to work nicely with natbib. 

\bibliographystyle{plainnat}
\bibliography{references}

\begin{thebibliography}{26}
\providecommand{\natexlab}[1]{#1}
\providecommand{\url}[1]{\texttt{#1}}
\expandafter\ifx\csname urlstyle\endcsname\relax
  \providecommand{\doi}[1]{doi: #1}\else
  \providecommand{\doi}{doi: \begingroup \urlstyle{rm}\Url}\fi

\bibitem[Aeronautiques et~al.(1998)Aeronautiques, Howe, Knoblock, McDermott, Ram, Veloso, Weld, SRI, Barrett, Christianson, et~al.]{aeronautiques1998pddl}
Constructions Aeronautiques, Adele Howe, Craig Knoblock, ISI~Drew McDermott, Ashwin Ram, Manuela Veloso, Daniel Weld, David~Wilkins SRI, Anthony Barrett, Dave Christianson, et~al.
\newblock Pddl| the planning domain definition language.
\newblock \emph{Technical Report, Tech. Rep.}, 1998.

\bibitem[Ahn et~al.(2022)Ahn, Brohan, Brown, Chebotar, Cortes, David, Finn, Fu, Gopalakrishnan, Hausman, et~al.]{ahn2022can}
Michael Ahn, Anthony Brohan, Noah Brown, Yevgen Chebotar, Omar Cortes, Byron David, Chelsea Finn, Chuyuan Fu, Keerthana Gopalakrishnan, Karol Hausman, et~al.
\newblock Do as i can, not as i say: Grounding language in robotic affordances.
\newblock \emph{arXiv preprint arXiv:2204.01691}, 2022.

\bibitem[Ding et~al.(2023)Ding, Zhang, Amiri, Cao, Yang, Kaminski, Esselink, and Zhang]{ding2023integrating}
Yan Ding, Xiaohan Zhang, Saeid Amiri, Nieqing Cao, Hao Yang, Andy Kaminski, Chad Esselink, and Shiqi Zhang.
\newblock Integrating action knowledge and llms for task planning and situation handling in open worlds.
\newblock \emph{arXiv preprint arXiv:2305.17590}, 2023.

\bibitem[Fox and Long(2003)]{fox2003pddl2}
Maria Fox and Derek Long.
\newblock Pddl2. 1: An extension to pddl for expressing temporal planning domains.
\newblock \emph{Journal of artificial intelligence research}, 20:\penalty0 61--124, 2003.

\bibitem[Gragera et~al.(2023)Gragera, Fuentetaja, Garc{\'\i}a-Olaya, and Fern{\'a}ndez]{gragera2023planning}
Alba Gragera, Raquel Fuentetaja, {\'A}ngel Garc{\'\i}a-Olaya, and Fernando Fern{\'a}ndez.
\newblock A planning approach to repair domains with incomplete action effects.
\newblock In \emph{Proceedings of the International Conference on Automated Planning and Scheduling}, volume~33, pages 153--161, 2023.

\bibitem[Helmert(2006)]{helmert2006fast}
Malte Helmert.
\newblock The fast downward planning system.
\newblock \emph{Journal of Artificial Intelligence Research}, 26:\penalty0 191--246, 2006.

\bibitem[Hoffmann(2003)]{hoffmann2003metric}
J{\"o}rg Hoffmann.
\newblock The metric-ff planning system: Translating``ignoring delete lists''to numeric state variables.
\newblock \emph{Journal of artificial intelligence research}, 20:\penalty0 291--341, 2003.

\bibitem[Huang et~al.(2022{\natexlab{a}})Huang, Abbeel, Pathak, and Mordatch]{huang2022language}
Wenlong Huang, Pieter Abbeel, Deepak Pathak, and Igor Mordatch.
\newblock Language models as zero-shot planners: Extracting actionable knowledge for embodied agents.
\newblock In \emph{International Conference on Machine Learning}, pages 9118--9147. PMLR, 2022{\natexlab{a}}.

\bibitem[Huang et~al.(2022{\natexlab{b}})Huang, Xia, Xiao, Chan, Liang, Florence, Zeng, Tompson, Mordatch, Chebotar, et~al.]{huang2022inner}
Wenlong Huang, Fei Xia, Ted Xiao, Harris Chan, Jacky Liang, Pete Florence, Andy Zeng, Jonathan Tompson, Igor Mordatch, Yevgen Chebotar, et~al.
\newblock Inner monologue: Embodied reasoning through planning with language models.
\newblock \emph{arXiv preprint arXiv:2207.05608}, 2022{\natexlab{b}}.

\bibitem[James et~al.(2019)James, Freese, and Davison]{james2019pyrep}
Stephen James, Marc Freese, and Andrew~J Davison.
\newblock Pyrep: Bringing v-rep to deep robot learning.
\newblock \emph{arXiv preprint arXiv:1906.11176}, 2019.

\bibitem[Li et~al.(2022)Li, Li, Xiong, and Hoi]{li2022blip}
Junnan Li, Dongxu Li, Caiming Xiong, and Steven Hoi.
\newblock Blip: Bootstrapping language-image pre-training for unified vision-language understanding and generation.
\newblock In \emph{International Conference on Machine Learning}, pages 12888--12900. PMLR, 2022.

\bibitem[Liang et~al.(2023)Liang, Huang, Xia, Xu, Hausman, Ichter, Florence, and Zeng]{liang2023code}
Jacky Liang, Wenlong Huang, Fei Xia, Peng Xu, Karol Hausman, Brian Ichter, Pete Florence, and Andy Zeng.
\newblock Code as policies: Language model programs for embodied control.
\newblock In \emph{2023 IEEE International Conference on Robotics and Automation (ICRA)}, pages 9493--9500. IEEE, 2023.

\bibitem[Lin et~al.(2021)Lin, Bercher, et~al.]{lin2021change}
Songtuan Lin, Pascal Bercher, et~al.
\newblock Change the world-how hard can that be? on the computational complexity of fixing planning models.
\newblock In \emph{IJCAI}, pages 4152--4159, 2021.

\bibitem[Lin et~al.(2023)Lin, Grastien, and Bercher]{lin2023towards}
Songtuan Lin, Alban Grastien, and Pascal Bercher.
\newblock Towards automated modeling assistance: An efficient approach for repairing flawed planning domains.
\newblock In \emph{Proceedings of the AAAI Conference on Artificial Intelligence}, volume~37, pages 12022--12031, 2023.

\bibitem[Liu et~al.(2023{\natexlab{a}})Liu, Jiang, Zhang, Liu, Zhang, Biswas, and Stone]{liu2023llm+}
Bo~Liu, Yuqian Jiang, Xiaohan Zhang, Qiang Liu, Shiqi Zhang, Joydeep Biswas, and Peter Stone.
\newblock Llm+ p: Empowering large language models with optimal planning proficiency.
\newblock \emph{arXiv preprint arXiv:2304.11477}, 2023{\natexlab{a}}.

\bibitem[Liu et~al.(2023{\natexlab{b}})Liu, Bahety, and Song]{liu2023reflect}
Zeyi Liu, Arpit Bahety, and Shuran Song.
\newblock Reflect: Summarizing robot experiences for failure explanation and correction.
\newblock \emph{arXiv preprint arXiv:2306.15724}, 2023{\natexlab{b}}.

\bibitem[Mishra et~al.(2022)Mishra, Khashabi, Baral, and Hajishirzi]{mishra2022cross}
Swaroop Mishra, Daniel Khashabi, Chitta Baral, and Hannaneh Hajishirzi.
\newblock Cross-task generalization via natural language crowdsourcing instructions.
\newblock In \emph{Proceedings of the 60th Annual Meeting of the Association for Computational Linguistics (Volume 1: Long Papers)}, pages 3470--3487, 2022.

\bibitem[OpenAI(2023)]{gpt4v}
OpenAI.
\newblock Gpt-4v(ision) system card.
\newblock \url{https://cdn.openai.com/papers/GPTV_System_Card.pdf}, 2023.

\bibitem[Ouyang et~al.(2022)Ouyang, Wu, Jiang, Almeida, Wainwright, Mishkin, Zhang, Agarwal, Slama, Ray, et~al.]{ouyang2022training}
Long Ouyang, Jeffrey Wu, Xu~Jiang, Diogo Almeida, Carroll Wainwright, Pamela Mishkin, Chong Zhang, Sandhini Agarwal, Katarina Slama, Alex Ray, et~al.
\newblock Training language models to follow instructions with human feedback.
\newblock \emph{Advances in Neural Information Processing Systems}, 35:\penalty0 27730--27744, 2022.

\bibitem[Radford et~al.(2021)Radford, Kim, Hallacy, Ramesh, Goh, Agarwal, Sastry, Askell, Mishkin, Clark, et~al.]{radford2021learning}
Alec Radford, Jong~Wook Kim, Chris Hallacy, Aditya Ramesh, Gabriel Goh, Sandhini Agarwal, Girish Sastry, Amanda Askell, Pamela Mishkin, Jack Clark, et~al.
\newblock Learning transferable visual models from natural language supervision.
\newblock In \emph{International conference on machine learning}, pages 8748--8763. PMLR, 2021.

\bibitem[Sanh et~al.(2022)Sanh, Webson, Raffel, Bach, Sutawika, Alyafeai, Chaffin, Stiegler, Le~Scao, Raja, et~al.]{sanh2022multitask}
Victor Sanh, Albert Webson, Colin Raffel, Stephen~H Bach, Lintang Sutawika, Zaid Alyafeai, Antoine Chaffin, Arnaud Stiegler, Teven Le~Scao, Arun Raja, et~al.
\newblock Multitask prompted training enables zero-shot task generalization.
\newblock In \emph{ICLR 2022-Tenth International Conference on Learning Representations}, 2022.

\bibitem[Singh et~al.(2023)Singh, Blukis, Mousavian, Goyal, Xu, Tremblay, Fox, Thomason, and Garg]{singh2023progprompt}
Ishika Singh, Valts Blukis, Arsalan Mousavian, Ankit Goyal, Danfei Xu, Jonathan Tremblay, Dieter Fox, Jesse Thomason, and Animesh Garg.
\newblock Progprompt: Generating situated robot task plans using large language models.
\newblock In \emph{2023 IEEE International Conference on Robotics and Automation (ICRA)}, pages 11523--11530. IEEE, 2023.

\bibitem[Sreedharan et~al.(2019)Sreedharan, Srivastava, Smith, and Kambhampati]{sreedharan2019can}
Sarath Sreedharan, Siddharth Srivastava, David Smith, and Subbarao Kambhampati.
\newblock Why can't you do that hal? explaining unsolvability of planning tasks.
\newblock In \emph{International Joint Conference on Artificial Intelligence}, 2019.

\bibitem[Valmeekam et~al.(2023)Valmeekam, Marquez, Olmo, Sreedharan, and Kambhampati]{valmeekam2023planbench}
Karthik Valmeekam, Matthew Marquez, Alberto Olmo, Sarath Sreedharan, and Subbarao Kambhampati.
\newblock Planbench: An extensible benchmark for evaluating large language models on planning and reasoning about change.
\newblock In \emph{Thirty-seventh Conference on Neural Information Processing Systems Datasets and Benchmarks Track}, 2023.

\bibitem[Wang et~al.(2023)Wang, Gonzalez-Pumariega, Sharma, and Choudhury]{wang2023demo2code}
Huaxiaoyue Wang, Gonzalo Gonzalez-Pumariega, Yash Sharma, and Sanjiban Choudhury.
\newblock Demo2code: From summarizing demonstrations to synthesizing code via extended chain-of-thought.
\newblock \emph{arXiv preprint arXiv:2305.16744}, 2023.

\bibitem[Xie et~al.(2023)Xie, Yu, Zhu, Bai, Gong, and Soh]{xie2023translating}
Yaqi Xie, Chen Yu, Tongyao Zhu, Jinbin Bai, Ze~Gong, and Harold Soh.
\newblock Translating natural language to planning goals with large-language models.
\newblock \emph{arXiv preprint arXiv:2302.05128}, 2023.

\end{thebibliography}

\clearpage
\appendix

\subsection{Experiment Details}
\subsubsection{Task}
The descriptions of the tasks used for evaluation are shown in Table \ref{tab_task_description}.

\begin{table}[h]
    \centering
    \begin{tabular}{|l|l|}
    \hline
    \textbf{Task}  &  \textbf{Task description} \\ 
    \hline
    serve fruit & Place a fruit on the plate. \\
    \hline
    store fruit & Store an apple in a closed fridge. \\
    \hline
    heat sandwich & Heat the sandwich and place it on a plate. \\
    \hline
    heat milk & Heat the milk and place it on the table.\\
    \hline
    \end{tabular}
    \caption{Tasks for evaluation.}
    \label{tab_task_description}
\end{table}

\subsubsection{Available Objects in the Environment}

We designed two environments for these four tasks. In the environment, the objects are divided into three groups: (1) receptacles equipped with doors, capable of containing items; (2) liquids; and (3) other objects.
The available objects in the environments are shown in Table \ref{objects}.

\begin{table}[h]
    \centering
    \begin{tabular}{|p{1.2cm}|p{3.2cm}|p{3.2cm}|}
    \hline
    \textbf{Env}  &  \textit{Env1} & \textit{Env2} \\ 
    \hline
    \textbf{receptacles} & microwave, fridge & microwave, fridge \\
    \hline
    \textbf{liquids} & water & milk \\
    \hline
    \textbf{objects} & glove, cup, plate, table, apple, watermelon, egg, beef & glove, cup, plate, table, sandwich, bottle, knife \\
    \hline
    \end{tabular}
    \caption{Avaliable objects in the environments for evaluation.}
    \label{objects}
\end{table}

\textit{Env1} is designed for the tasks of \textit{serving fruit} and \textit{storing fruit}, and \textit{Env2} is tailored for the tasks of \textit{heating sandwich}  and \textit{heating milk}.

\subsection{Additional Experiment}

COWP~\cite{ding2023integrating} is an open-world task planning method which also combines LLMs and symbolic planners. COWP relies on the provided situation to determine which action would encounter an error. In contrast, our proposed method can determine the cause of an error based on textual observations of the execution. Therefore, COWP relies on more prior information than our method. 

Here, we compare our method with COWP and its variant: COWP uses the ground-truth cause of the error as input, while COWP-O feeds the textual observation of the execution error to COWP. Since COWP can currently only generate the preconditions involving binary-valued predicates and deal with only one situation, we used test cases 2, 4, 5, and 6 to evaluate the performance of COWP and its variant. The result is shown in Table \ref{tab_cowp_results}. Note that COWP requires a provided cause of error to achieve performance similar to our method. However, given the textual observations used by our method, COWP does not perform as well. Moreover, the reason for COWP-C achieving a 0\% success rate on test case 4 is that the semantics of the generated predicate (e.g., ``\textit{microwave\_is\_open}") resemble an existing predicate, ``\textit{is\_microwave}", rather than the correct predicate, ``\textit{is\_open}". This mismatch leads to the generated predicate aligning with an incorrect value.

\begin{table}[h]
    \centering
    \begin{tabular}{c|c|c|c}
    \toprule
    % \hline
        Case & COWP~\cite{ding2023integrating} & COWP-O  & LASP (ours)  \\
        \hline
        2 & 100\% & 40\%  & 100\% \\
        4 & 0\% & 0\% & 100\% \\
        5 & 80\% & 40\% & 80\%\\
        6 & 100\% & 40\%  & 100\% \\
    \bottomrule
    % \hline
    \end{tabular}
    \caption{Comparisons with COWP and its variant.}
    \label{tab_cowp_results}
\end{table}

\subsection{Prompts for Baselines}

\subsubsection{Language Models as Zero-Shot Planners (LMZSP)\cite{huang2022language}}

The prompt we used to implement LMZSP is shown as follows:

\promtblock{\underline{\textbf{Prompt:}}
Task: place the apple on the table.\newline
Step 1: open fridge\newline
Step 2: pick apple\newline
Step 3: place apple on table\newline
\\
Task: heat the soup.\newline
Step 1: open microwave\newline
Step 2: open fridge\newline
Step 3: pick bowl\newline
Step 4: place bowl in microwave\newline
Step 5: close microwave\newline
Step 6: heat soup in bowl with microwave\newline
\blank{[\$TASK DESCRIPTION\$]} Task: heat the sandwich and place it on a plate\\
\\
\llmoutput{\underline{\textbf{LLM output:}}
Step 1: open microwave\\
Step 2: pick up sandwich\\
Step 3: place sandwich in microwave\\
Step 4: close microwave\\
Step 5: heat sandwich in microwave\\
Step 6: open cupboard\\
Step 7: pick up plate\\
Step 8: remove heated sandwich from microwave using oven mitts\\
Step 9: place hot sandwich on plate
}}

\subsubsection{Inner Monologue~\cite{huang2022inner}}
The prompts utilized for implementing Inner Monologue consist of two components. The first involves querying the LLM to generate a plan, while the second entails querying the LLM to adjust the plan in response to feedback. These two types of prompts are outlined below:

\promtblock{\underline{\textbf{Prompt:}}
You are a household robot. A human user can require you to do various tasks. You should break down the task into the sequence of actions that you can perform to accomplish the task.\\
\blank{[\$PRIMITIVE ACTIONS\$]}: You can perform the following actions: pick object, place object on object, place object on receptacle, place object in receptacle, open receptacle, close receptacle, heat object with receptacle, heat liquid in object with receptacle, wear heat-insulator, pour liquid from object to object.\\
And you need to specify the object, receptacle, or liquid for each action. \newline
Human: place the apple on the table.\\
Scene: [ objects: apple, table; receptacles: fridge; liquids: .]\\
Robot: \\
1. open fridge\\
2. pick apple\\
3. place apple on table}
\promtblock{Human: heat the soup.\\
Scene: [ objects: potato, plate, bowl, table; receptacles: microwave, fridge; liquids: soup.]\\
Robot: \\
1. open microwave\\
2. open fridge\\
3. pick bowl\\
4. place bowl in microwave\\
5. close microwave\\
6. heat soup in bowl with microwave\\
\blank{[\$TASK DESCRIPTION\$]}: Human: heat the sandwich and place it on a plate.\\
Scene: [ objects: \blank{[\$OBJECTS\$]}; receptacles: \blank{[\$RECEPTACLES\$]}; liquids: \blank{[\$LIQUIDS\$]}. ]
\newline
\newline
\llmoutput{\underline{\textbf{LLM output:}} Robot:\\
1. open microwave\\
2. pick sandwich\\
3. place sandwich in microwave\\
4. close microwave\\
5. heat sandwich with microwave\\
6. open microwave\\
7. wear heat-insulator\\
8. pick heated sandwich\\
9. pick plate\\
10. place heated sandwich on plate}}

\promtblock{\underline{\textbf{Feedback prompt:}}
\blank{[\$TRANSLATED PLAN\$]}: The actual plan executed by the robot:\\
1. open microwave\\
2. pick sandwich\\
3. place sandwich in microwave\\
4. close microwave\\
5. heat sandwich with microwave\\
6. open microwave\\
7. wear heat-insulator\\
8. pick heated sandwich\\
9. pick plate\\
10. place sandwich on plate\\
Human: The robot encountered an error when executing the \blank{[\$ERRONEOUS ACTION INDEX\$]} (e.g., third) step of the plan. \blank{[\$ERROR OBSERVATION\$]}: The observation of the error is: The robot struggled to open the microwave while holding an object, causing it to accidentally drop the object.\\
\\
\llmoutput{\underline{\textbf{LLM output:}} Robot: \\
1. open fridge\\
2. pick sandwich\\
3. place sandwich on table\\
4. open microwave\\
5. pick sandwich from table\\
6. place sandwich in microwave\\
7. close microwave\\
8. heat sandwich with microwave\\
9. open microwave\\
10. wear heat-insulator\\
11. pick heated sandwich\\
12. pick plate\\
13. place heated sandwich on plate}}

Since the output of the LLM may not follow the format of the primitve actions, we follow LMZSP \cite{huang2022language} to use a pre-trained language model to translate each output action into an available action. \blank{[\$TRANSLATED PLAN\$]} is the result of the translation.

\begin{figure}[t!]
\centering
\includegraphics[width=0.48\textwidth]{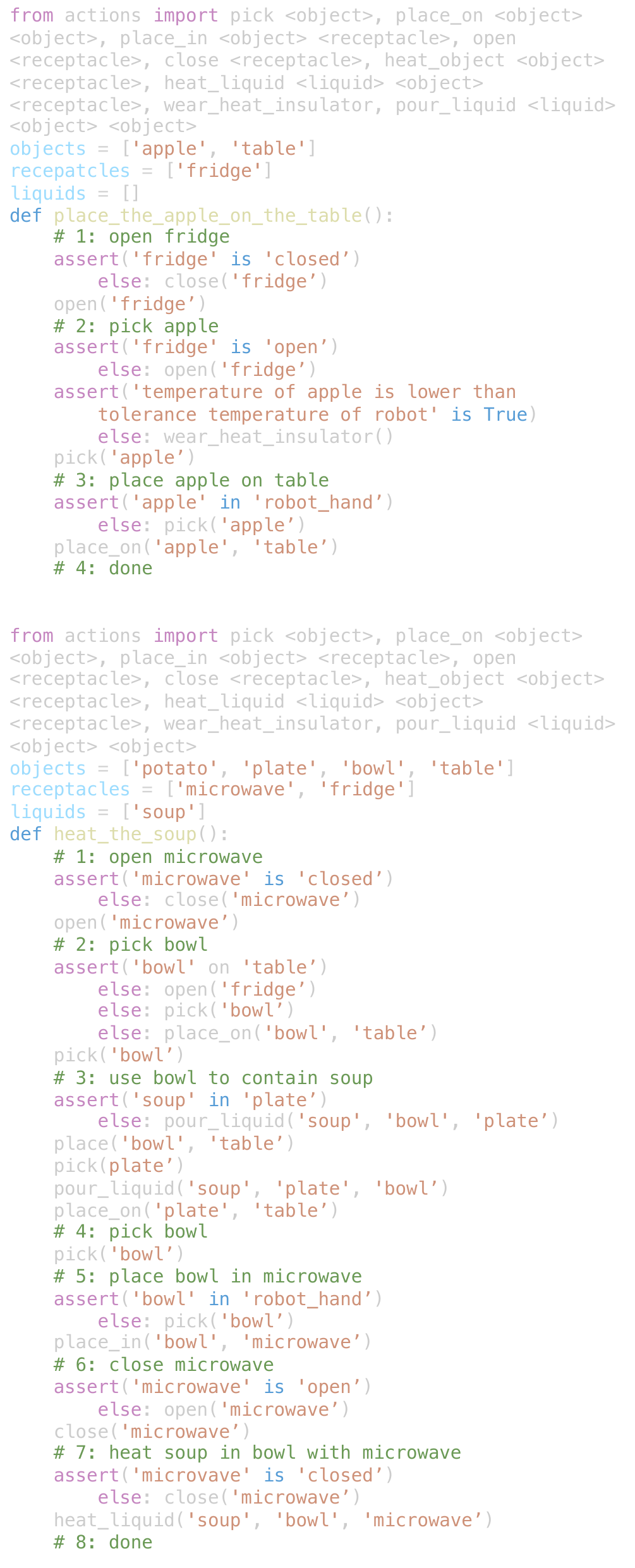}
\caption{Two code examples used in ProgPrompt.} \label{codes}
\end{figure}

\subsubsection{ProgPrompt \cite{singh2023progprompt}}
\begin{figure}[h]
\centering
\includegraphics[width=0.48\textwidth]{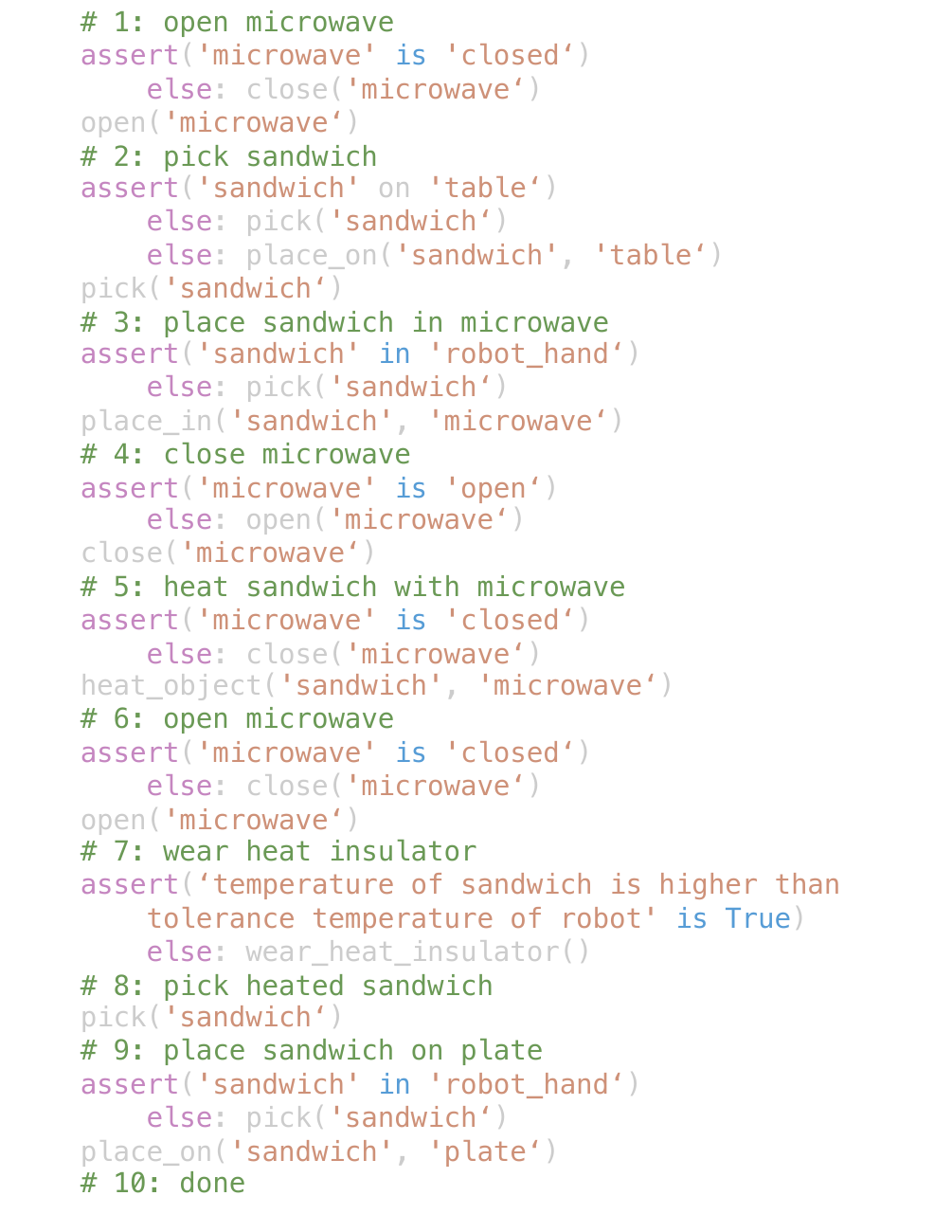}
\caption{A code example generated by ProgPrompt.} \label{llm_out_code}
\end{figure}

The prompt utilized for implementing ProgPrompt is shown below:

\promtblock{\underline{\textbf{Prompt:}} You are a code expert. Given the available functions, parameters, and the function name, you are required to implement the function. You need to follow the grammar shown in the following examples. Here are two examples:\\
\textcolor{light-blue}{[\$TWO CODE EXAMPLES\$]}: See Fig.\ref{codes}\\
Given \blank{[\$CODE-STYLE PRIMITIVE ACTIONS\$]} \blank{[\$CODE-STYLE OBJECTS\$]} \blank{[\$CODE-STYLE RECEPTACLES\$]} \blank{[\$CODE-STYLE LIQUIDS\$]}\\
\blank{[\$CODE-STYLE TASK DESCRIPTION\$]}: \\def heat\_the\_sandwich\_and\_place\_it\_on\_a\_plate():
\\
\\
\underline{\textbf{LLM output:}} See Fig. \ref{llm_out_code}} 
where \blank{[\$CODE-STYLE PRIMITIVE ACTIONS\$]}, \blank{[\$CODE-STYLE OBJECTS\$]}, \blank{[\$CODE-STYLE RECEPTACLES\$]}, and \blank{[\$CODE-STYLE LIQUIDS\$]} represent primitive actions, objects, receptacles, and liquids in the specified code style, respectively. Some examples are shown in Fig.\ref{codes}.

\subsection{Conversations of LLM in LASP}
Here, we present two records of the input and output of the LLM during the runtime of the LASP algorithm. In one instance, LASP successfully augmented the planner's knowledge, resulting in a correct plan. Conversely, in the other instance, it encountered failure. In each instance, we present the input and output of five modules: (1) cause analyzer, (2) action precondition generator, (3) property completeness evaluator, (4) NL-to-PDDL translator, and (5) object expander.

The record of the successful case is shown below:

(1) Cause analyzer:

\promtblock{\underline{\textbf{Prompt:}} You are expected to provide possible reasons for errors in robot actions, where errors are essentially unexpected outcomes. You are given the robot task, the historical actions, the action caused the error, the observation of the error. In your answer, each reason should be as brief as possible. And the answer should follow this format:\\
Possible reasons:\\
1. reason1\\
2. reason2\\
 ...\\
The robot task is to heat the sandwich and place it on a plate.\\
The historical actions are:\\
1. Open the fridge.\\
2. Open the microwave.\\
3. Pick up the sandwich from the fridge.\\
4. Put the sandwich in the microwave.\\
5. Close the microwave.\\
6. Use the microwave to heat the sandwich.\\
7. Open the microwave.\\
The action caused the error is: Pick up the sandwich from the microwave.}
\promtblock{The observation of the error is: The agent's sensors detected excessive heat and immediately released the sandwich.\\
What are possible reasons that could lead to this error?}
\promtblock{\llmoutput{\underline{\textbf{LLM output:}} Possible reasons:\\
1. The robot's gripping mechanism may be faulty or not designed to withstand high temperatures.\\
2. There may be a fault in heat detection sensor causing it to overreact in normal circumstances.\\
3. The microwave could have overheated the sandwich beyond usual temperature.\\
4. The robot may not have waited for the sandwich to cool down slightly before attempting to pick it up.\\
5. The robot's programming may not include protocols for handling hot objects.\\
6. Perhaps the design of the robot does not take into account the variable heat of objects coming out of a microwave.}}

(2) Action precondition generator:

\promtblock{\underline{\textbf{Prompt:}} You are given a robotic action that caused an error and the reason why this error occurred. You are expected to provide a precondition for an robotic action to prevent a recurrence of the same error. The precondition should only require one property of the object or robot satisfy a certain condition. The answer should be as brief as possible (no further explanation) and it should follow this format:\\
Precondition: ...\\
The action caused the error is: Pick up the sandwich from the microwave.\\
The reason why this error occurred is: The robot's gripping mechanism may be faulty or not designed to withstand high temperatures.\\
What precondition should be satisfied to prevent a recurrence of the same error?}
\promtblock{\llmoutput{\underline{\textbf{LLM output:}} Precondition: The robot's gripping mechanism must be heat-resistant and in good working order.}}

(3) Property completeness evaluator:

\promtblock{\underline{\textbf{Prompt:}} You are expected to judge whether the given natural language described precondition can be expressed by the candidate predicates and functions, or their respective antonyms. You are given the natural language described precondition, candidate predicates and functions. Answer should be ``Yes" or ``No". If you answer ``Yes", you are expected to provide a suitable predicate or two suitable functions for expressing the precondition. The suitable predicate or two suitable functions must come from the given candidate predicates and functions. There are two kinds of formats to provide the suitable expression: (1) If suggesting a predicate, respond with: ``Suitable predicate: ..."; (2) If suggesting two functions, respond with: ``Suitable functions: ...".\\
Natural language described precondition: The robot's gripping mechanism must be heat-resistant and in good working order.\\
Candidate predicates: (holding ?r - robot ?obj - object), (is-empty-handed ?r - robot), (is-open ?rec - receptacle), (in ?obj - object ?rec - receptacle), (on ?obj - object ?loc - location), (is-microwave ?rec - receptacle), (is-heat-insulation ?obj - object), (liquid-in ?liq - liquid ?obj - object), (can-contain-liquid ?obj - object), (is-microwave-safe ?obj - object), (can-support ?loc - location).\\
Candidate functions: (temperature ?obj - object), (tolerance-temperature ?r - robot), (liquid-temperature ?liq - liquid).\\
Can the precondition be expressed by the candidate predicates and functions, or their respective antonyms? If yes, provide a suitable predicate or two suitable functions for expressing the precondition.\\
\\
\llmoutput{\underline{\textbf{LLM output:}} Yes\\
Suitable functions: (temperature ?obj - object), (tolerance-temperature ?r - robot)}}

(4) NL-to-PDDL translator:

\promtblock{\underline{\textbf{Prompt:}} Given the action description, the candidate parameters, the corresponding between instances and parameters, candidate functions, and a natural utterance, you are expected to translate the natural utterance into a PDDL precondition expression for the given action. The expected PDDL expression should satisfy the following grammar: ($<$comparison-operator$>$ $<$function-1$>$ $<$function-2$>$). The grammar of $<$function$>$ is ($<$function-name$>$ $<$parameter-1$>$ $<$parameter-2$>$ ...). $<$function-name$>$ must come from the given candidate functions. $<$parameter$>$ must be one of the candidate parameters. $<$comparsion-operator$>$ should be one of the following: ``$<$", ``$>$", ``$<=$", ``$>=$". The answer should follow this format (no further explanation):\\
PDDL expression: ...\\
Here is an example:\\
Action description: Pick up the box from the cabinet.\\
Candidate parameters: the robot, denoted as $?$r; the object, denoted as $?$obj; the receptacle, denoted as $?$rec.\\
Corresponding between instances and parameters: the robot is $?$r; the box is $?$obj; the cabinet is $?$rec.\\
Candidate functions: (weight $?$obj), (lift-capacity $?$r)\\
Natural utterance: The weight of the box should not exceed the maximum weight that the robot can handle.\\
PDDL expression: ($<=$ (weight $?$obj) (lift-capacity $?$r))}
\promtblock{Action description: Pick up the sandwich from the microwave.\\
Candidate parameters: the robot, denoted as $?$r; the object, denoted as $?$obj; the receptacle, denoted as $?$rec.\\
Corresponding between instances and parameters: the agent is $?$r; the sandwich is $?$obj; the microwave is $?$rec.\\
Candidate functions: (temperature $?$obj - object), (tolerance-temperature $?$r - robot).\\
Natural utterance: The robot's gripping mechanism must be heat-resistant and in good working order.\\
\\
\llmoutput{\underline{\textbf{LLM output:}} PDDL expression: ($<=$ (temperature $?$obj) (tolerance-temperature $?$r))}}

(5) Object expander:

\promtblock{\underline{\textbf{Prompt:}} You are expected to select the most suitable object from the candidate objects to prevent a recurrence of the error and assist the robot complete the task. You are given the robot task, the action caused the error, the precondition that the action does not satisfy, and the candidate objects. Answer should follow this format: Suitable object: $<$object$>$. $<$object$>$ must be one of the candidate objects.\\
The robot task is to heat the sandwich and place it on a plate.\\
The action caused the error is: Pick up the sandwich from the microwave.\\
The precondition that the action does not satisfy is: The robot's gripping mechanism must be heat-resistant and in good working order.\\
Candidate objects: cup, glove, knife.\\
Which object is the most suitable to prevent a recurrence of the error and assist the robot complete the task?\\
\\
\llmoutput{\underline{\textbf{LLM output:}} Suitable object: glove.}}

The record of the failed case is shown below:

% \paragraph{haha}
(1) Cause analyzer:

\promtblock{\underline{\textbf{Prompt:}} You are expected to provide possible reasons for errors in robot actions, where errors are essentially unexpected outcomes. You are given the robot task, the historical actions, the action caused the error, the observation of the error. In your answer, each reason should be as brief as possible. And the answer should follow this format:\\
Possible reasons:\\
1. reason1\\
2. reason2\\
 ...\\
The robot task is to heat the milk and place it on the table.\\
The historical actions are:\\
1. Open the fridge.\\
2. Pick up the sandwich from the fridge.\\
3. Put the sandwich on the table.\\
4. Pick up the bottle from the fridge.\\
The action caused the error is: Pour the milk from the bottle into the sandwich which is on the table.\\
The observation of the error is: The milk permeated the sandwich, soaking it, and some overflowed onto the table.}
\promtblock{What are possible reasons that could lead to this error?\\
\\
\llmoutput{\underline{\textbf{LLM output:}} Possible reasons:\\
1. Misinterpretation of instructions.\\
2. Inability to differentiate between objects.\\
3. Programming error in the sequence of actions.\\
4. Faulty sensor causing incorrect object identification.\\
5. Error in algorithm guiding robot's decision-making process.\\
6. Skipped action or incorrect execution order.\\
7. Potential bottle manipulation or handling issues.}}

In the failed case, LASP did not correctly augment the planner's knowledge based on all the possible reasons shown above. Here, we use the third reason as an example to show the input and output of the subsequent  modules.
\\

(2) Action precondition generator:

\promtblock{\underline{\textbf{Prompt:}} You are given a robotic action that caused an error and the reason why this error occurred. You are expected to provide a precondition for an robotic action to prevent a recurrence of the same error. The precondition should only require one property of the object or robot satisfy a certain condition. The answer should be as brief as possible (no further explanation) and it should follow this format:\\
Precondition: ...\\
The action caused the error is: Pour the milk from the bottle into the sandwich which is on the table.\\
The reason why this error occurred is: Programming error in the sequence of actions.\\
What precondition should be satisfied to prevent a recurrence of the same error?\\
\\
\llmoutput{\underline{\textbf{LLM output:}} Precondition: The target object for pouring must be a container.}}

(3) Property completeness evaluator:

\promtblock{\underline{\textbf{Prompt:}} You are expected to judge whether the given natural language described precondition can be expressed by the candidate predicates and functions, or their respective antonyms. You are given the natural language described precondition, candidate predicates and functions. Answer should be ``Yes" or ``No". If you answer ``Yes", you are expected to provide a suitable predicate or two suitable functions for expressing the precondition. The suitable predicate or two suitable functions must come from the given candidate predicates and functions. There are two kinds of formats to provide the suitable expression: (1) If suggesting a predicate, respond with: ``Suitable predicate: ..."; (2) If suggesting two functions, respond with: ``Suitable functions: ...".\\
Natural language described precondition: The target object for pouring must be a container.\\
Candidate predicates: (holding ?r - robot ?obj - object), (is-empty-handed ?r - robot), (is-open ?rec - receptacle), (in ?obj - object ?rec - receptacle), (on ?obj - object ?loc - location), (is-microwave ?rec - receptacle), (is-heat-insulation ?obj - object), (liquid-in ?liq - liquid ?obj - object), (is-microwave-safe ?obj - object), (can-support ?loc - location).\\
Candidate functions: (temperature ?obj - object), (tolerance-temperature ?r - robot), (liquid-temperature ?liq - liquid).}
\promtblock{Can the precondition be expressed by the candidate predicates and functions, or their respective antonyms? If yes, provide a suitable predicate or two suitable functions for expressing the precondition.\\
\\
\llmoutput{\underline{\textbf{LLM output:}} Yes\\
Suitable predicate: (is-open ?rec - receptacle)}}

(4) NL-to-PDDL translator:

\promtblock{\underline{\textbf{Prompt:}} Given the action description, the candidate parameters, the corresponding between instances and parameters, candidate predicate, and a natural utterance, you are expected to translate the natural utterance into a PDDL precondition expression for the given action.\\
The expected PDDL expression should satisfy one of following grammars: (1) ($<$predicate$>$ $<$parameter-1$>$ $<$parameter-2$>$ ...); (2) (not ($<$predicate$>$ $<$parameter-1$>$ $<$parameter-2$>$ ...)). $<$parameter$>$ must be one of the candidate parameters. $<$predicate$>$ should be the given candidate predicate. The answer should follow this format (no further explanation):\\
PDDL expression: ...\\
Here is an example:\\
Action description: Put the plate in the fridge.\\
Candidate parameters: the robot, denoted as ?r; the object, denoted as ?obj; the receptacle, denoted as ?rec.\\
Corresponding between instances and parameters: the robot is ?r; the plate is ?obj; the fridge is ?rec.\\
Candidate predicate: (is-open ?rec) \\
Natural utterance: The fridge should be open when placing the plate in the fridge.\\
PDDL expression: (is-open ?rec)\\
Action description: Pour the milk from the bottle into the sandwich which is on the table.\\
Candidate parameters: the robot, denoted as ?r; the liquid, denoted as ?liq; the object, denoted as ?obj1; the object, denoted as ?obj2; the location, denoted as ?loc.\\
Corresponding between instances and parameters: the agent is ?r; the milk is ?liq; the bottle is ?obj1; the sandwich is ?obj2; the table is ?loc.\\
Candidate predicate: (is-open ?rec - receptacle).\\
Natural utterance: The target object for pouring must be a container.\\
\\
\llmoutput{\underline{\textbf{LLM output:}} PDDL expression: (is-open ?obj2)}}

In this case, the property completeness evaluator falsely recognized the precondition described in natural language can be expressed by the existing predicate. Consequently, the NL-to-PDDL translator translated it into an improper PDDL statement. Besides, the predicate (is-open ?obj - object) is not present in the PDDL domain file, yet LASP inferred that the precondition can be expressed by existing predicates. In such a scenario, LASP will proceed to explore the next reason.

\end{document}